\documentclass[sigconf]{acmart}

\usepackage{balance}

\newcommand{\eref}[1]{Eqn.~\eqref{#1}}  
\newcommand{\figref}[1]{Fig.~\ref{#1}}  
\newcommand{\prg}[1]{\noindent\textbf{#1}.} 
\usepackage{tabularx}
\usepackage{multirow}
\usepackage[font=small,labelfont=bf]{caption}
\usepackage{bigstrut}
\usepackage{multirow}
\usepackage{hyperref}
\hypersetup{
    colorlinks=true,
    linkcolor=blue,
    urlcolor=blue,
}
\usepackage{titlesec}
\titlespacing*{\subsection}{0pt}{0.15\baselineskip}{0.05\baselineskip}
\titlespacing*{\section}{0pt}{0.6\baselineskip}{0.5\baselineskip}

\AtBeginDocument{%
  \providecommand\BibTeX{{%
    \normalfont B\kern-0.5em{\scshape i\kern-0.25em b}\kern-0.8em\TeX}}}

\copyrightyear{2020}
\acmYear{2020}
\setcopyright{acmcopyright}
\acmConference[HRI '20]{Proceedings of the 2020 ACM/IEEE International Conference on
Human-Robot Interaction}{March 23--26, 2020}{Cambridge, United Kingdom}
\acmBooktitle{Proceedings of the 2020 ACM/IEEE International Conference on Human-Robot
Interaction (HRI '20), March 23--26, 2020, Cambridge, United Kingdom}
\acmPrice{15.00}
\acmDOI{10.1145/3319502.3374832}
\acmISBN{978-1-4503-6746-2/20/03}
\usepackage{xcolor}

\begin{document}
\fancyhead{} 
\title{When Humans Aren't Optimal: \\ Robots that Collaborate with Risk-Aware Humans}
\author{Minae Kwon$^*$, Erdem Biyik$^*$, Aditi Talati$^\dagger$, Karan Bhasin$\ddagger$, Dylan P. Losey$^*$, Dorsa Sadigh$^*$}
\affiliation{%
  \institution{$^*$ Stanford University, $\dagger$ American High School, $\ddagger$ The Harker School}
  \city{[mnkwon, ebiyik, dlosey, dorsa]@stanford.edu, talatiaditi02@gmail.com,  karanbhasin03@gmail.com}
}

\begin{abstract}
    
    In order to collaborate safely and efficiently, robots need to anticipate how their human partners will behave. Some of today's robots model humans as if they were also robots, and assume users are always \textit{optimal}. Other robots account for human limitations, and relax this assumption so that the human is \textit{noisily} rational. Both of these models make sense when the human receives deterministic rewards: i.e., gaining either $\$100$ or $\$130$ with certainty. But in real-world scenarios, rewards are rarely deterministic. Instead, we must make choices subject to \textit{risk} and \textit{uncertainty}---and in these settings, humans exhibit a cognitive bias towards \textit{suboptimal} behavior. For example, when deciding between gaining $\$100$ with certainty or $\$130$ only $80\%$ of the time, people tend to make the risk-averse choice---even though it leads to a lower expected gain! In this paper, we adopt a well-known \textit{Risk-Aware} human model from behavioral economics called Cumulative Prospect Theory and enable robots to leverage this model during human-robot interaction (HRI). In our user studies, we offer supporting evidence that the Risk-Aware model more accurately predicts suboptimal human behavior. We find that this increased modeling accuracy results in safer and more efficient human-robot collaboration. Overall, we extend existing rational human models so that collaborative robots can anticipate and plan around suboptimal human behavior during HRI.
    
    
\end{abstract}





\maketitle
\begin{figure}[t h!]
	\begin{center}
		\includegraphics[scale=0.18]{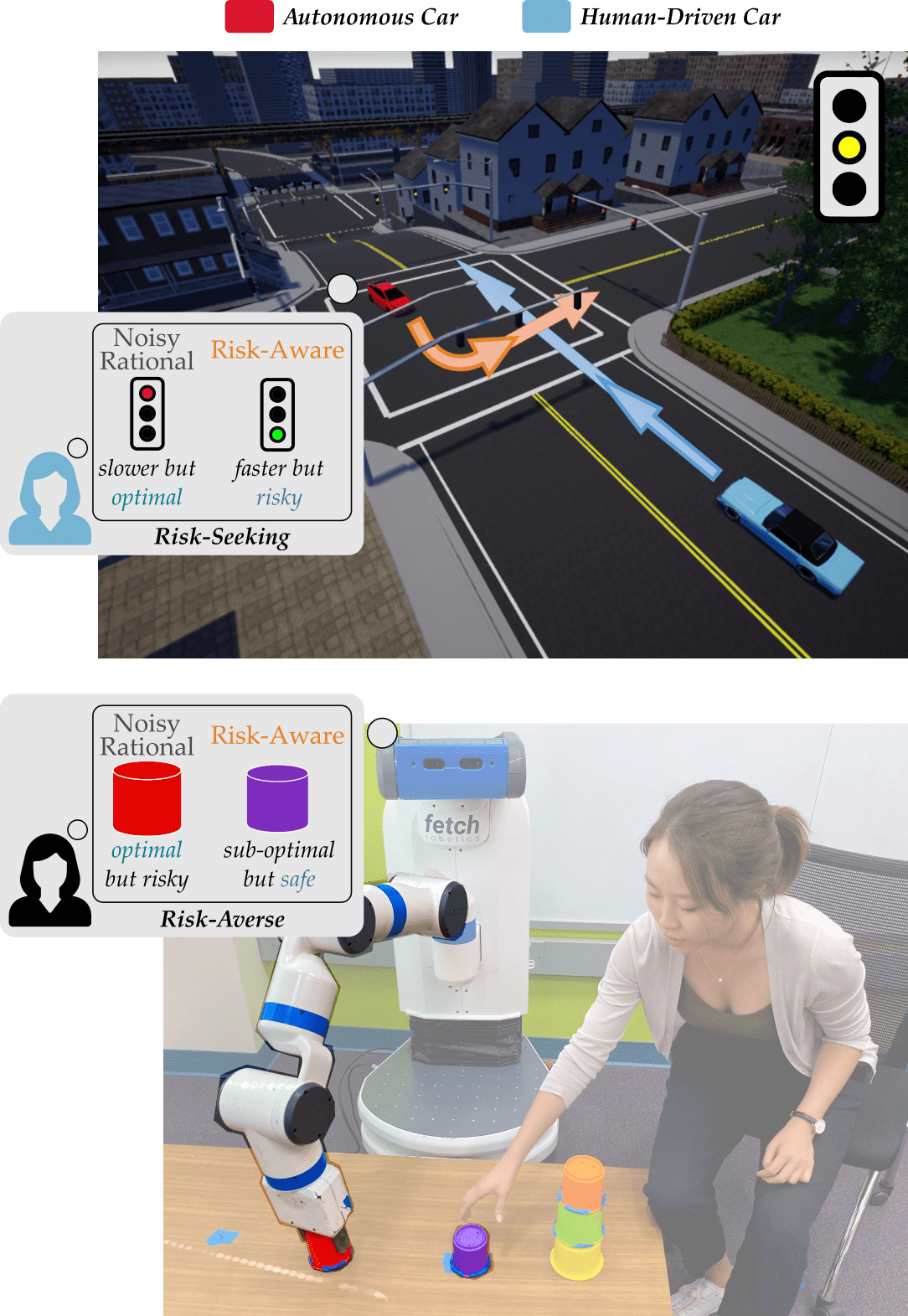}

		\caption{\small{Robots collaborating with humans that must deal with risk. (Top) Autonomous car predicting if the human will try to make the light. (Bottom) Robot arm anticipating which cup the human will grab. In real world scenarios, people exhibit a cognitive bias towards irrational but Risk-Aware behavior.}}
		\label{fig:front}
	\end{center}
	\vspace{-1em}
\end{figure}

\section{Introduction}

When robots collaborate with humans, they must \emph{anticipate} how the human will behave for seamless and safe interaction. Consider a scenario where an autonomous car is waiting at an intersection (see top of Fig.~\ref{fig:front}). The autonomous car wants to make an unprotected left turn, but a human driven car is approaching in the oncoming lane. The human's traffic light is yellow, and will soon turn red. Should the autonomous car predict that this human will stop---so that the autonomous car can safely turn left---or anticipate that the human will try and make the light---where turning left leads to a collision?

Previous robots anticipated that humans acted like robots, and made \emph{rational} decisions to maximize their reward \cite{gray2013robust, raman2015reactive, vitus2013probabilistic, vasudevan2012safe, ng2000algorithms, abbeel2004apprenticeship}. However, assuming humans are always rational fails to account for the limited time, computational resources, and noise that affect human decision making, and so today's robots anticipate that humans make \emph{noisily} rational choices \cite{dragan2013legibility,sadigh2017active,ziebart2008maximum,finn2016guided,palan2019learning}. Under this model, the human is always most likely to choose the action leading to the highest reward, but the robot also recognizes that the human may behave suboptimally. This makes sense when humans are faced with deterministic rewards: e.g., the light will definitely turn red in $5$ seconds. Here, the human knows whether or not they will make the light, and can accelerate or decelerate accordingly. But in real world settings, we usually do not have access to deterministic rewards. Instead, we need to deal with uncertainty and estimate \emph{risk} in every scenario. Returning to our example, imagine that the human has a $95\%$ chance of making the light if they accelerate: success saves some time during their commute, while failure could result in a ticket or even a collision. It is still rational for the human to decelerate; however, a risk-seeking user will attempt to make the light. How the robot models the human affects the \emph{safety} and \emph{efficiency} of this interaction: a Noisy Rational robot believes it should turn left, while a \emph{Risk-Aware} robot realizes that the human is likely to run the light, and waits to prevent a collision.

When robots treat nearby humans as noisily rational, they \emph{miss out} on how risk biases human decisions. Instead, we assert
\vspace{-3px}
\begin{center}
    \emph{To ensure safe and efficient interaction, robots must recognize\\ that people behave suboptimally when risk is involved.}
\end{center}
\vspace{-3px}
Our approach is inspired by behavioral economics, where results indicate that users maintain a nonlinear transformation between \emph{actual} and \emph{perceived} rewards and probabilities \cite{kahneman2013prospect, tversky1992advances}. 
Here, the human over- or under-weights differences between rewards, resulting in a cognitive bias (a systematic error in judgment) that leads to risk-averse or risk-seeking behavior. We equip robots with this cognitive model, enabling them to anticipate risk-affected human behavior and better collaborate with humans in everyday scenarios (see \figref{fig:front}).

Overall, we make the following contributions:

\noindent\textbf{Incorporating Risk in Robot Models of Humans.}
We propose using Cumulative Prospect Theory as a Risk-Aware model. We formalize a theory-of-mind (ToM) where the robot models the human as reacting to their decisions or environmental conditions. We integrate Cumulative Prospect Theory into this formalism so that the robot can model suboptimal human actions under risk. 
In a simulated autonomous driving environment, our user studies demonstrate that the Risk-Aware robot more accurately predicts the human's behavior than a Noisy Rational baseline. \\
\noindent\textbf{Determining when to Reason about Risk.} We identify the types of scenarios where reasoning about risk is important. Our results suggest that scenarios with close expected rewards is the most important in determining whether humans will act suboptimally.\\
\noindent\textbf{Safe and Efficient Collaboration when Humans face Uncertainty.}
We develop planning algorithms so that robots can leverage our Risk-Aware human model to improve collaboration. In a collaborative cup stacking task, shown on the bottom in \figref{fig:front}, the Risk-Aware robotic arm anticipated that participants would choose suboptimal but risk-averse actions, and planned trajectories to avoid interfering with the human's motions. Users completed the task more efficiently with the Risk-Aware robot, and also subjectively preferred working with the Risk-Aware robot over the Noisy Rational baseline.

This work describes a computationally efficient and empirically supported way for robots to model suboptimal human behavior by extending the state-of-the-art to also account for risk. A summary of our paper, including videos of experiments, can be found \href{https://www.youtube.com/watch?v=PnBNI1ms0iw}{\textit{here}}.
\section{Related Work}
Previous work has shown that robots that successfully predict humans' behavior exhibit improved performance in many applications, such as assistive robotics \cite{losey2019controlling,awais2010human,dragan2012formalizing,losey2019enabling}, motion planning \cite{ziebart2009planning,nikolaidis2017human}, collaborative games \cite{nguyen2011capir}, and autonomous driving \cite{bai2015intention,sadigh2016planning,sadigh2016information}. One reason behind this success is that human modeling equips robots with a theory of mind (ToM), or the ability to attribute a mind to oneself and others \cite{premack1978does,thomaz2016computational}. \citet{devin2016implemented} showed ToM can improve performance in human-robot collaboration.


For this purpose, researchers have developed various human models. In robotics, the Noisy Rational choice model has remained extremely popular due to its simplicity. Several works in reward learning \cite{sadigh2017active,biyik2019active,biyik2019green,basu2019active,palan2019learning,brown2019ranking}, reinforcement learning \cite{finn2016guided}, inverse reinforcement learning \cite{ramachandran2007bayesian,ziebart2008maximum,bloem2014infinite}, inverse planning \cite{baker2009action}, and human-robot collaboration \cite{pellegrinelli2016human} employed the noisy rational model for human decision-making. Other works developed more complex human models and methods specifically for autonomous driving \cite{liebner2012driver,sadigh2016information,vasudevan2012safe,gray2013robust}. Unfortunately, these models either assume humans are rational or do not handle situations with uncertainty and risk. There have been other human models that take a learning-based approach ~\cite{osogami2014restricted,otsuka2016deep, unhelkar2019learning}. While this is an interesting direction, these methods are usually not very data efficient.

In cognitive science, psychology and behavioral economics, researchers have developed other decision-making models. For example, \citet{ordonez1997decisions} investigated decision making under time constraints; \citet{diederich1997dynamic} developed a model based on stochastic processes to model humans' process of making a selection between two options, again under a time constraint. \citet{ortega2016human} proposed a rationality model based on concepts from information theory and statistical mechanics to model time-constrained decision making. \citet{mishra2014decision} studied decision making under risk from the perspectives of biology, psychology and economics. \citet{halpern2014decision} modeled the humans as a finite automata, and \citet{simon1972theories} developed \emph{bounded rationality} to incorporate suboptimalities and constraints. \citet{evans2016learning} investigated different biases humans can have in decision-making. Among all of these works, Cumulative Prospect Theory (CPT) \cite{tversky1992advances} remains prominent as it successfully models suboptimal human decision making under risk. Later works studied how Cumulative Prospect Theory can be employed for time-constrained decision making \cite{young2012decision,eilertsen2014cumulative}.

In this paper, we adopt Cumulative Prospect Theory as an example of a Risk-Aware model. We show that it not only leads to more accurate predictions of human actions, but also increases the performance of the robot and the human-robot team.


\section{Formalism}\label{sec:formalism}
We assume a setting where a human needs to select from a set of actions  $\mathcal{A}_H$. Each action $a_H\in\mathcal{A}_H$ may have several possible consequences, where, without loss of generality, we denote the number of consequences as $K$. For a given human action $a_H$, we express the probabilities of each consequence and their corresponding rewards as a set of pairs: 
\begin{align*}
   C(a_H) =\left\{\left(p^{(1)}, R^{(1)}_H(a_H)\right), \left(p^{(2)}, R^{(2)}_H(a_H)\right), \dots, \left(p^{(K)}, R^{(K)}_H(a_H)\right) \right\} 
\end{align*}

We outline and compare two methods that use $C(a_H)$ to model human actions: Noisy Rational and Cumulative Prospect Theory (CPT) ~\cite{tversky1992advances}. CPT is a prominent model of human decision-making under risk \cite{young2012decision,eilertsen2014cumulative} and we use it as an example of a Risk-Aware model. Finally, we describe how we can integrate them into a partially observable Markov decision process (POMDP) formulation of human-robot interaction. 

\noindent\textbf{Noisy Rational Model.} According to the noisy rational model, humans are more likely to choose actions with the highest expected reward, and are less likely to choose suboptimal actions (i.e., they are optimal with some noise). The noise comes from constraints such as limited time or computational resources. For instance, in the autonomous driving example, Noisy Rational model would predict the human will most likely choose the optimal action and decelerate. Denoting the expected reward of the human for action $a_H$ as
\begin{align*}
    R_H(a_H) = p^{(1)}R^{(1)}_H(a_H) + p^{(2)}R^{(2)}_H(a_H) + \dots, p^{(K)}R^{(K)}_H(a_H),
\end{align*}
the noisy rational model asserts
\begin{align}
    P(a_H) = \frac{\exp{\left(\theta \cdot R_H(a_H)\right)}}{\sum_{a\in\mathcal{A}_H}\exp{\left(\theta \cdot R_H(a)\right)}},
    \label{eq:noisy_rational}
\end{align}
where $\theta \in [0, \infty)$ is a temperature parameter, commonly referred to as the \emph{rationality coefficient}, which controls how noisy the human is. While larger $\theta$ models the human as a better reward maximizer, setting $\theta\! =\! 0$ means the human chooses actions uniformly at random.

Hence, the Noisy Rational model is simply a linear transformation of the reward with the rationality coefficient $\theta\geq0$, which makes the transformation monotonically non-decreasing. As the model does not transform the probability values, it becomes impossible to model suboptimal humans using this approach. The closest Noisy Rational can get to modeling suboptimal humans is to assign a uniform probability to all actions.

\noindent\textbf{Risk-Aware Model.} We adopt Cumulative Prospect Theory (CPT) \cite{tversky1992advances} as an example of a Risk-Aware model. According to this model, humans are not simply Noisy Rational. They may, for example, be suboptimally risk-seeking or risk-averse. For instance, in the autonomous driving example, human drivers can be risk-seeking and try to make the yellow light even though they risk a costly collision. The Risk-Aware model captures suboptimal decision-making by transforming both the probabilities and the rewards. These transformations aim to represent what humans actually perceive. 
The reward transformation is a pairwise function:
\begin{align*}
    v(R) = \begin{cases}
        R^\alpha \textrm{ if } R\geq0\\
        -\lambda(-R)^\beta \textrm{ if } R<0\\
    \end{cases}.
\end{align*}
The parameters $\alpha, \beta\in[0,1]$ represent how differences among rewards are perceived. For instance, when $\alpha,\beta\in(0,1)$, the model predicts that humans will perceive differences between large positive (or negative) rewards as relatively lower than the differences between smaller positive (resp. negative) rewards, even though the true differences are equal. $\lambda\in[0,\infty)$ characterizes how much more (or less) important negative rewards are compared to positive rewards. When $\lambda>1$, humans are modeled as loss-averse, assigning more importance to losses compared to gains. The reverse is true when $\lambda \in [0,1)$. 


The Risk-Aware model also implements a transformation over the probabilities. The probabilities $(p^{(1)},p^{(2)},\dots)$ are divided into two groups based on whether their corresponding true rewards are positive or negative. The probability transformations corresponding to positive and negative rewards ($w^+, w^-$) are as follows:
\begin{align*}
    w^+(p) = \frac{p^\gamma}{(p^\gamma + (1-p)^\gamma)^{1/\gamma}}, \; w^-(p) = \frac{p^\delta}{(p^\delta + (1-p)^\delta)^{1/\delta}},
\end{align*}
where $\gamma,\delta\in[0,1]$.

Without loss of generality, we assume that each of the $K$ rewards are ordered in decreasing order, i.e. $R^{(i+1)}_H(a)\leq R^{(i)}_H(a)$ for all $i\in\{1,2,\dots,K-1\}$ and $a\in\mathcal{A}_H$. Then, the probability transformation is as follows:
\begin{align*}
    \pi\left(C(a_H)\right) &= \left(\pi^+(C(a_H)),\pi^-(C(a_H))\right)\\
    \pi^+(C(a_H)) &= \left(w^+\left(p^{(1)}\right),w^+\left(p^{(1)}+p^{(2)}\right)-w^+\left(p^{(1)}\right),\dots\right)\\
    \pi^-(C(a_H)) &=
    \left(\dots,w^-\left(p^{(K)}\!+\!p^{(K-1)}\right)\!-\!w^-\left(p^{(K)}\right),w^-\left(p^{(K)}\right)\right)
\end{align*}
Finally, we normalize probabilities so that $\pi(C(a_H))$ sums to 1: 
\begin{align*}
    \overline{\pi}_j(C(a_H)) = \frac{\pi_j(C(a_H))}{\sum_{i=1}^{K} \pi_i(C(a_H)))}, \forall j\in\{1,2,\dots,K\}
\end{align*}

When $\gamma, \delta \in (0,1)$, the probability transformations capture biases humans are reported to have ~\cite{tversky1992advances} by overweighting smaller probabilities and underweighting larger probabilities. 

Based on these two transformations, we now extend the human decision making model with the Risk-Aware model:
\begin{align}
R^{\textrm{CPT}}_H(a_H) \!=\! \overline{\pi}_1(C(a_H))&\cdot v\left(R^{(1)}_H(a_H)\right) +\! \dots \!+ \overline{\pi}_K(C(a_H))
\cdot v\left(R^{(K)}_H(a_H)\right) \nonumber\\
P(a_H) &= \frac{\exp{\left(\theta \cdot R^{\textrm{CPT}}_H(a_H)\right)}}{\sum_{a\in\mathcal{A}_H}\exp{\left(\theta \cdot R^{\textrm{CPT}}_H(a)\right)}}\;.
\label{eq:risk_aware}
\end{align}


In contrast to the Noisy Rational model, the Risk-Aware model's expressiveness allows it to model \emph{both} optimal and suboptimal human decisions by assigning larger likelihoods to those actions.

\prg{Formal Model of Interaction}
We model the world where both the human and the robot take actions as a POMDP, which we denote with a tuple $\langle\mathcal{S},\mathcal{O},O, \mathcal{A}_H,\mathcal{A}_R,T,r_H,r_R\rangle$. $\mathcal{S}$ is the finite set of states; $\mathcal{O}$ is the set of observations; $O:\mathcal{S}\to\mathcal{O}$ defines the shared observation mapping; $\mathcal{A}_H$ and $\mathcal{A}_R$ are the finite action sets for the human and the robot, respectively; $T:\mathcal{S}\times\mathcal{A}_H\times\mathcal{A}_R\times\mathcal{S}\to [0,1]$ is the transition distribution. $r_H$ and $r_R$ are the reward functions that depend on the state, the actions and the next state. In this POMDP, we assume the agents act simultaneously. 
Having a first-order ToM, the human tries to optimize her own cumulative reward given an action distribution for the robot, $P(a_R | s)$. The human value function $V_H(s)$ can then be defined using the following Bellman update: 
\begin{align*}
    &V_H(s) = \max_{a_H\in\mathcal{A}_H} \mathbb{E}_{a_R|s}\mathbb{E}_{s'\mid s, a_H, a_R}\left[R_H(s,a_H,a_R,s') + \gamma V_H(s')\right]\;.
\end{align*}
We then use the fact that
\begin{align*}
    P(s,s',a_R \mid o, a_H) = P(s\mid o)\cdot P(a_R \mid s)\cdot P(s' \mid s, a_H, a_R)
\end{align*}
to construct a set $C(a_H)$ for the current observation $o$ that consists of the pairs $(P(s, s',a_R \mid o, a_H), V_H(s'))$ for varying $s$, $s'$, and $a_R$. When modeling the human as zeroth-order ToM, $P(a_R \mid s)$ will simply be a uniform distribution.

Having constructed $C(a_H)$, we can define the human's utility function for different values of $s$, and $s'$. The utility functions for both Noisy Rational and Risk-Aware models are defined as follows: 

\noindent \textbf{Noisy Rational}:
\begin{align*}
    R_H(o,a_H) = \sum_{s,s'\in\mathcal{S}}\sum_{a_R\in\mathcal{A}_R} P(s,s',a_R \mid o, a_H)\cdot V_H(s')\:.
\end{align*}

\noindent \textbf{Risk-Aware}:
\begin{align*}
    R_H^{CPT}(o,a_H)= \sum_{s,s'\in\mathcal{S}}\sum_{a_R\in\mathcal{A}_R} \bar{\pi}_i(P(s,s',a_R \mid o, a_H))\cdot v\left(V_H(s')\right)\;,
\end{align*}
where the index $i$ corresponds to the event that leads to $s'$ from $s$ with $a_R$, $a_H$. An optimal human would always pick the action $a_H$ that maximizes $\mathbb{E}_{s\mid o}\mathbb{E}_{a_R\mid s}\mathbb{E}_{s'\mid s,a_H,a_R}\left[V_H(s')\right]$. 
The robot can obtain $P(a_H|o)$ using \eref{eq:noisy_rational}, \eref{eq:risk_aware} and use it to maximize its own cumulative reward. 

\smallskip

\prg{Summary} We have outlined two ways in which we can model humans (\textit{Noisy Rational} and \textit{Risk-Aware}), and how we can formalize these models in a human-robot interaction setting. In the following section, we empirically analyze factors that allow the Risk-Aware robot to more accurately model human actions.
\section{Autonomous Driving}
\label{sec:driving}

\begin{table}[t]

    \caption{Autonomous Driving. Users were given different amounts of \textit{information} about the likelihood that the light would turn red. Under \textit{risk}, we list two tested probabilities of the light turning red.}
    \label{tab:params}
    \vspace{-1em}
    \begin{center}
    \begin{tabular}{{l}{l}}
        \textbf{\textit{Information}} & \textbf{\textit{Time}} \bigstrut[b]  \\ \hline
        \textbf{None}: \textit{With some probability} & \textbf{Timed}: $8$ s   \bigstrut[t] \\
        \textit{\quad the light will turn red.} & \textbf{Not Timed}: no limit \\
        
        \textbf{Explicit}: \textit{There is a 5\% chance} & \bigstrut[t] \\
        \textit{\quad the light will turn red.} & \textbf{\textit{Risk}} \\
        
        \textbf{Implicit}: \textit{Of the previous 380 cars} & \textbf{High}: $95\%$ \bigstrut[t] \\
        \textit{\quad that decided to accelerate,} & \textbf{Low}: $5\%$ \\
        \textit{\quad the light turned red for 19 cars.} & \bigstrut[b] \\
        \hline
    \end{tabular}
    \vspace{-10px}
    \end{center}
\end{table}

In our first user study, we focus on the autonomous driving scenario from the bottom of \figref{fig:front}. Here the autonomous car---which wants to make an unprotected left turn---needs to determine whether the human-driven car is going to try to make the light. We asked human drivers whether they would accelerate or stop in this scenario. Specifically, we adjusted the \textit{information} and \textit{time} available for the human driver to make their decision. We also varied the level of \textit{risk} by changing the probability that the light would turn red. Based on the participant's choices in each of these cases, we learned \textit{Noisy Rational} and \textit{Risk-Aware} human models. Our results demonstrate that autonomous cars that model humans as \textit{Risk-Aware} are better able to explain and anticipate the behavior of human drivers, particularly when drivers make suboptimal choices.

\smallskip

\noindent\textbf{Experimental Setup.} We used the driving example shown in \figref{fig:front}. Human drivers were told that they are returning a rental car, and are approaching a light that is currently yellow. If they run the red light, they have to pay a $\$500$ ticket. But stopping at the light will prevent the human from returning their rental car on time, which also has an associated fine! Accordingly, the human drivers had to decide between \textit{accelerating} (and potentially running the light) or \textit{stopping} (and returning the rental car with a fine).

\smallskip

\noindent\textbf{Independent Variables.} We varied the amount of \textit{information} and \textit{time} that the human drivers had to make their decision. We also tested two different \textit{risk} levels: one where accelerating was optimal, and one where stopping was optimal. Our parameters for information, time, and risk are provided in Table~\ref{tab:params}. \\
\noindent\textit{Information.} We varied the amount of information that the driver was given on three levels: None, Explicit, and Implicit. Under None, the driver must rely on their own prior to assess the probability that the light will turn red. By contrast, in Explicit we inform the driver of the exact probability. Because probabilities are rarely given to us in practice, we also tested Implicit, where drivers observed other peoples' experiences to estimate the probability of a red light.\\
\noindent\textit{Time.} We compared two levels for time: a Timed setting, where drivers had to make their choice in under $8$ seconds, and a Not Timed setting, where drivers could deliberate as long as necessary. \\
\noindent\textit{Risk.} We varied risk along two levels: High and Low. When the risk was High, the light turned red $95\%$ of the time, and when risk was Low, the light turned red only $5\%$ of the time.

\smallskip

\noindent\textbf{Participants and Procedure.} We conducted a within-subjects study on Amazon Mechanical Turk and recruited 30 participants. All participants had at least a 95\% approval rating and were from the United States. After providing informed consent, participants were first given a high-level description of the autonomous driving task and were shown the example from \figref{fig:front}. In subsequent questions, participants were asked to indicate whether they would accelerate or stop. We presented the Timed questions first and the Not Timed questions second. For each set of Timed and Not Timed questions, we presented questions in the order of their informativeness from None to Explicit. The risk levels were presented in random order \footnote{To learn more about our study, please check out our autonomous driving survey link: \url{https://stanfordgsb.qualtrics.com/jfe/form/SV_cUgxaZIvEkdb3ud}}. 
\smallskip

\begin{figure}[t]
	\begin{center}
		\includegraphics[width=1\columnwidth]{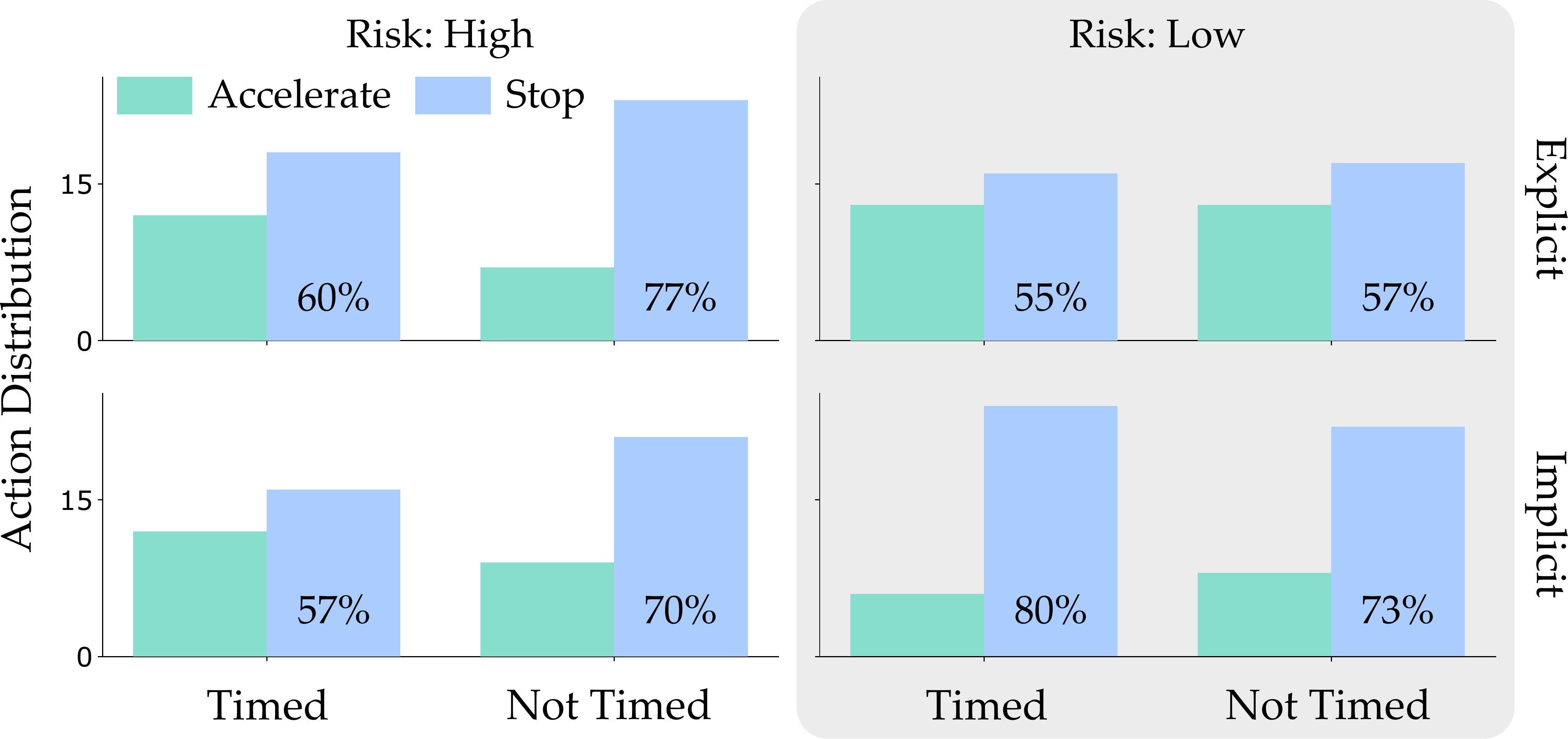}
		\caption{Action distributions for human drivers. Across all surveyed factors (information, time, and risk), more users preferred to stop at the light. Interestingly, stopping was the \textit{suboptimal} choice when the light rarely turns red (Low).}
		\label{fig:carlo_AD}
	\end{center}
	\vspace{-1em}
\end{figure}

\begin{figure*}[t]
    \includegraphics[scale=0.3]{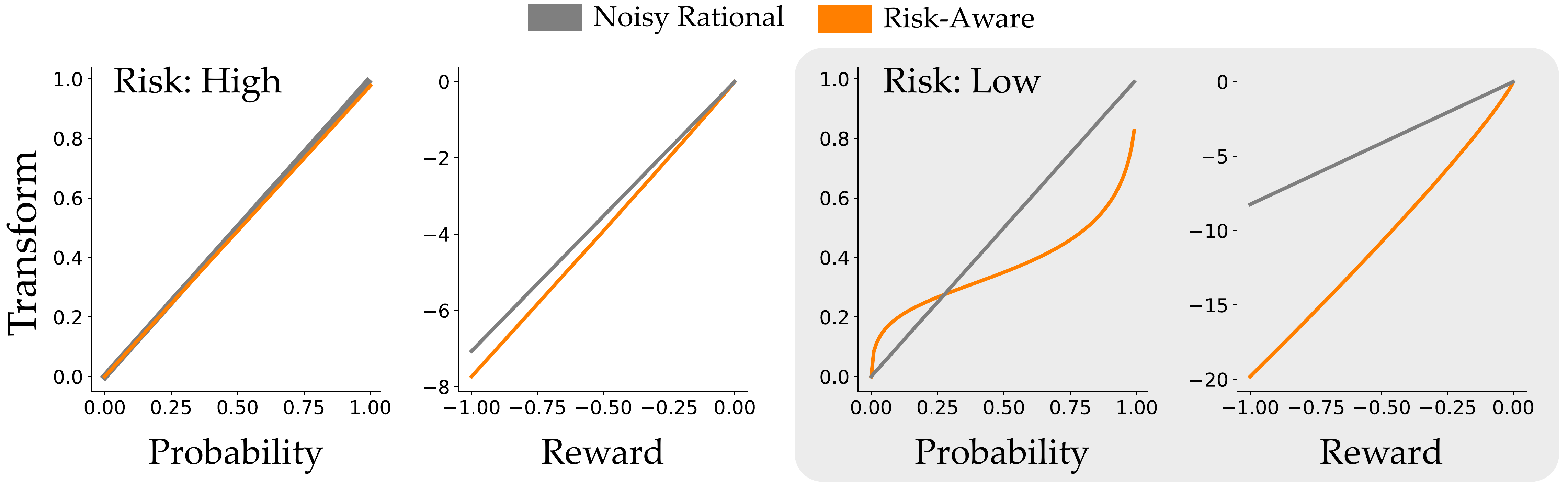}
    \caption{Averaged probability and reward transformations for human drivers that are modeled as \textit{Noisy Rational} or \textit{Risk-Aware}. In scenarios where the light frequently turns red (High), both models produce similar transformations. But when the light rarely turns red (Low), the models diverge: here the \textit{Risk-Aware} autonomous car recognizes that human drivers \textit{overestimate} both the probability that light will turn red and the cost of running the light. This enables \textit{Risk-Aware} autonomous cars to explain why human drivers prefer to stop, even though accelerating is the optimal action when the light rarely turns red.}
    \label{fig:carlo_TF}
\end{figure*}

\begin{figure}[t]
	\begin{center}
		\includegraphics[width=1\columnwidth]{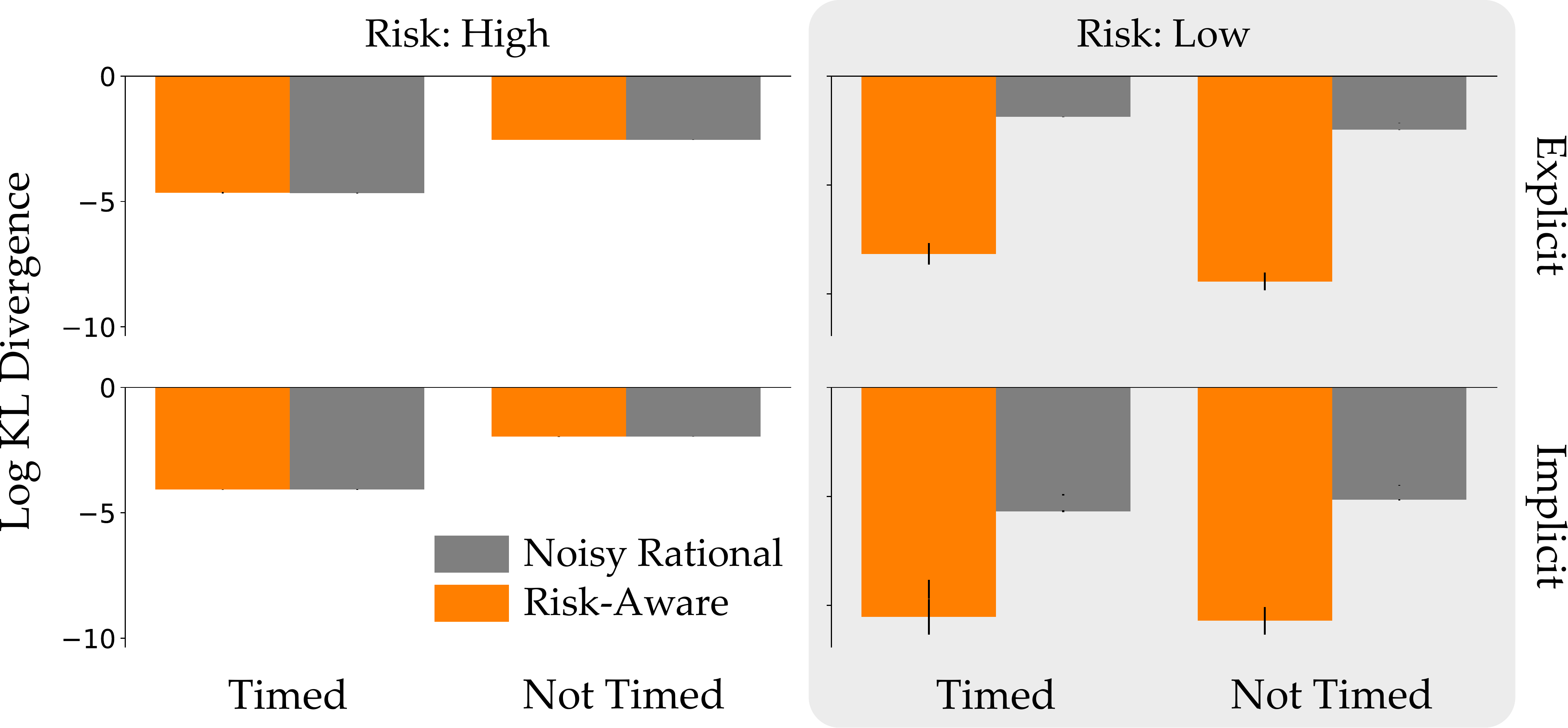}
		\caption{Model accuracy (lower is better). When the light often turns red (High), both models could anticipate the human's behavior. But when the light rarely turns red (Low), only the \textit{Risk-Aware} autonomous car correctly anticipated that the human would stop.}
		\label{fig:carlo_KL}
	\end{center}
	\vspace{-1em}
\end{figure}

\noindent\textbf{Dependent Measures.} We aggregated the user responses into \textit{action distributions}. These action distributions report the percentage of human drivers who chose to accelerate and stop under each treatment level. Next, we learned \textit{Noisy Rational} and \textit{Risk-Aware} models of human drivers for the autonomous car to leverage\footnote{To learn the models, we used the Metropolis-Hastings algorithm ~\cite{chib1995understanding} and obtained $30$ independent samples of the model parameters.}. To measure the accuracy of these models, we compared the Kullback-Leibler (KL) divergence between the \textit{true} action distribution and the model's \textit{predicted} action distribution. We report the \textit{log KL divergence} for both \textit{Noisy Rational} and \textit{Risk-Aware} models.

\smallskip

\noindent\textbf{Hypothesis.}
\begin{quote}
    \textbf{H1.} \textit{Autonomous cars which use \textit{Risk-Aware} models of human drivers will more accurately predict human action distributions than autonomous cars who treat humans as \textit{noisily rational} agents.}
\end{quote}

\noindent\textbf{Baseline.} In order to confirm that our users were trying to make optimal choices, we also queried the human drivers for their preferred actions in settings where the expected rewards were \textit{far apart} (e.g., where the expected reward for accelerating was much higher than the expected reward for stopping). In these baseline trials, users overwhelmingly chose the \textit{optimal} action ($93\%$ of trials).

\smallskip

\noindent\textbf{Results.} The results from our autonomous driving user study are summarized in Figs.~\ref{fig:carlo_AD}, \ref{fig:carlo_TF}, and \ref{fig:carlo_KL}. In each of the tested situations, most users elected to stop at the light (see \figref{fig:carlo_AD}). Although stopping at the light is the optimal action in the High risk case---where the light turns red $95\%$ of the time---stopping was actually \textit{suboptimal} in the Low risk case---where the light only turns red $5\%$ of the time. Because humans chose optimal actions in some cases (High risk) and suboptimal actions in other situations (Low risk), the autonomous car interacting with these human drivers must be able to anticipate \textit{both} optimal and suboptimal behavior.

In cases where the human was rational, autonomous cars learned similar \textit{Noisy Rational} and \textit{Risk-Aware} models (see \figref{fig:carlo_TF}). However, the \textit{Risk-Aware} model was noticeably different in situations where the human was suboptimal. Here autonomous cars using our formalism learned that human drivers \textit{overestimated} the likelihood that the light would turn red, and \textit{underestimated} the reward of running the light. Viewed together, the \textit{Risk-Aware} model suggests that human drivers were risk-averse when the light rarely turned red, and risk-neutral when the light frequently turned red.

Autonomous cars using our \textit{Risk-Averse} model of human drivers were better able to predict how humans would behave (see \figref{fig:carlo_KL}). Across all treatment levels, \textit{Risk-Averse} attained a log KL divergence of $-5.7 \pm 3.3$, while \textit{Noisy Rational} only reached $-3.3 \pm 1.3$. This difference was statistically significant ($t(239)=-11.5$, $p<.001$). Breaking our results down by risk, in the High case both models were similarly accurate, and any differences were insignificant ($t(119)=.42$, $p=.67$). But in the Low case---where human drivers were suboptimal---the \textit{Risk-Averse} model significantly outperformed the \textit{Noisy Rational} baseline ($t(119)=-17.3$, $p<.001$).

Overall, the results from our autonomous driving user study support hypothesis H1. Autonomous cars leveraging a \textit{Risk-Aware} model were able to understand and anticipate human drivers both in situations where the human is optimal \textit{or} suboptimal, while the \textit{Noisy Rational} model could not explain why the participants preferred to take a safer (but suboptimal) action.

\smallskip

\begin{figure}[t]
	\begin{center}
		\includegraphics[width=0.9\columnwidth]{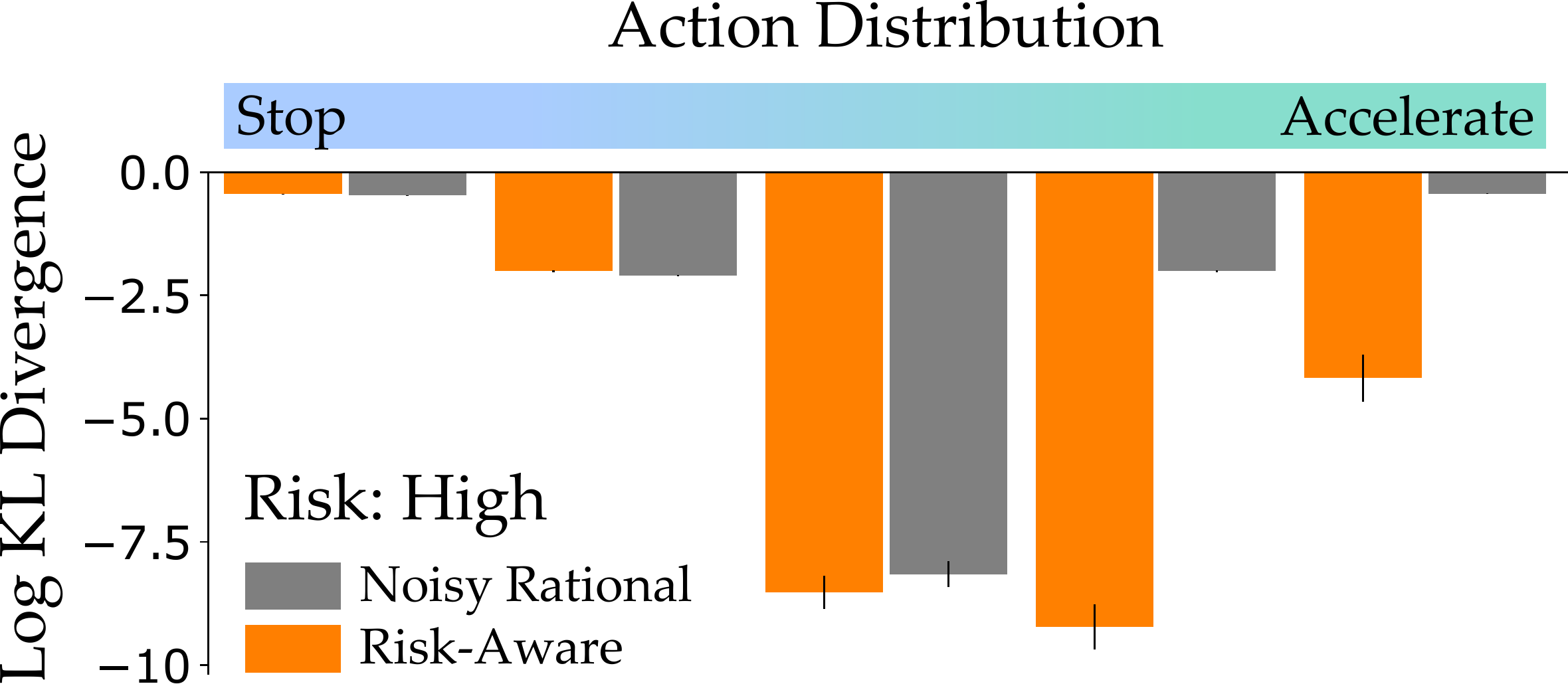}
		\caption{Model accuracy (lower is better) on simulated data. We simulated a spectrum of human action distributions in scenarios where the light often turns red (High risk). The optimal action here is to \textit{stop}. As the human becomes increasingly optimal, both \textit{Noisy Rational} and \textit{Risk-Aware} provide a similarly accurate prediction. But when the human is suboptimal---accelerating through the light---the \textit{Risk-Aware} autonomous car yields a more accurate prediction.}
		\label{fig:carlo_SIM}
	\end{center}
	\vspace{-1.5em}
\end{figure}

\noindent\textbf{Follow-up: Disentangling Risk and Suboptimal Decisions.} After completing our user study, we performed a simulated experiment within the autonomous driving domain. Within this experiment, we \textit{fixed} the probability that the light would turn red, and then \textit{varied} the human driver's action distribution. When fixing the probability, we used the High risk scenario where the optimal decision was to \textit{stop}. The purpose of this follow-up experiment was to make sure that our model can also explain suboptimally \textit{aggressive} drivers, and to ensure that our results are not tied to the Low risk scenario. Our simulated results are displayed in \figref{fig:carlo_SIM}. As before, when the human driver chose the optimal action, both \textit{Noisy Rational} and \textit{Risk-Aware} models were equally accurate. But when the human behaved aggressively---and tried to make the light---only the \textit{Risk-Aware} autonomous car could anticipate their suboptimal behavior. These results suggest that the improved accuracy of the \textit{Risk-Aware} model is tied to user \textit{suboptimality}, and not to the particular type of risk (either High or Low).

\smallskip

\prg{Summary} We find supporting evidence that \textit{Risk-Aware} is more accurate at modeling human drivers in scenarios that involve decision making under uncertainty. In particular, our results suggest that the reason why \textit{Risk-Aware} is more effective at modeling human drivers is because humans often act \textit{suboptimally} in these scenarios. When humans act rationally, both \textit{Noisy Rational} and \textit{Risk-Aware} autonomous cars can understand and anticipate their actions.
\section{Collaborative Cup Stacking}
\label{sec:stacking}

Within the autonomous driving user studies, we demonstrated that our \textit{Risk-Aware} model enables robots to \textit{accurately} anticipate their human partners. Next, we want to explore how our formalism leverages this accuracy to improve \textit{safety} and \textit{efficiency} during HRI. To test the usefulness of our model, we performed two user studies with a 7-DoF robotic arm (Fetch, Fetch Robotics). In an \textbf{online} user study, we verify that the \textit{Risk-Aware} model can accurately model humans in a collaborative setting. In an \textbf{in-person} user study, the robot leverages \textit{Risk-Aware} and \textit{Noisy Rational} models to anticipate human choices and plan trajectories that avoid interfering with the participant. Both studies share a common experimental setup, where the human and robot collaborate to stack cups into a tower.

\smallskip

\noindent\textbf{Experimental Setup.} The collaborative cup stacking task is shown in \figref{fig:front} (also see the supplemental video). We placed five cups on the table between the person and robot. The robot knew the location and size of the cups \textit{a priori}, and had learned motions to pick up and place these cups into a tower. However, the robot did not know which cups its human partner would pick up. 

The human chooses their cups with two potential towers in mind: an \textit{efficient but unstable} tower, which was more likely to fall, or a \textit{inefficient but stable} tower, which required more effort to assemble. Users were awarded $20$ points for building the stable tower (which never fell) and $105$ for building the unstable tower (which collapsed $80\%$ of the time). Because the expected utility of building the unstable tower was higher, our \textit{Noisy Rational} baseline anticipated that participants would make the unstable tower.

\smallskip

\noindent\textbf{Independent Variables.} We varied the robot's model of its human partner with two levels: \textit{Noisy Rational} and \textit{Risk-Aware}. The \textit{Risk-Aware} robot uses our formalism from Section~\ref{sec:formalism} to anticipate how humans make decisions under uncertainty and risk.

\smallskip

\subsection{Anticipating Collaborative Human Actions}

\smallskip

Our \textbf{online} user study extended the results from the autonomous driving domain to this collaborative cup stacking task. We focused on how \textit{accurately} the robot anticipated the participants' choices.

\smallskip

\noindent\textbf{Participants and Procedure.} We recruited 14 Stanford affiliates and 36 Amazon Mechanical Turkers for a total of 50 users (32\% Female, median age: 33). Participants from Amazon Mechanical Turk had at least a 95\% approval rating and were from the United States. After providing informed consent, each of our users answered survey questions about whether they would collaborate with the robot to build the \textit{efficient but unstable} tower, or the \textit{inefficient but stable} tower. Before users made their choice, we explicitly provided the rewards associated with each tower, and implicitly gave the probability of the tower collapsing. To implicitly convey the probabilities, we showed videos of humans working with the robot to make stable and unstable towers: all five videos with the stable tower showed successful trials, while only one of the five videos with the unstable tower displayed success. After watching these videos and considering the rewards, participants chose their preferred tower type \footnote{To learn more about our study please check out our cup stacking survey link: \url{https://stanfordgsb.qualtrics.com/jfe/form/SV_0oz1Y04mQ0s3i7P}}.

\smallskip

\noindent\textbf{Dependent Measures.} We aggregated the participants' decisions to find their \textit{action distribution} over stable and unstable towers. We fit \textit{Noisy Rational} and \textit{Risk-Aware} models to this action distribution, and reported the \textit{log KL divergence} between the actual tower choices and the choices predicted by the models.

\smallskip

\noindent\textbf{Hypotheses.}
\begin{quote}
    \textbf{H2.} \textit{Risk-Aware robots will better anticipate which tower the collaborative human user is attempting to build.}
\end{quote}

\begin{figure}[t]
	\begin{center}
		\includegraphics[width=1\columnwidth]{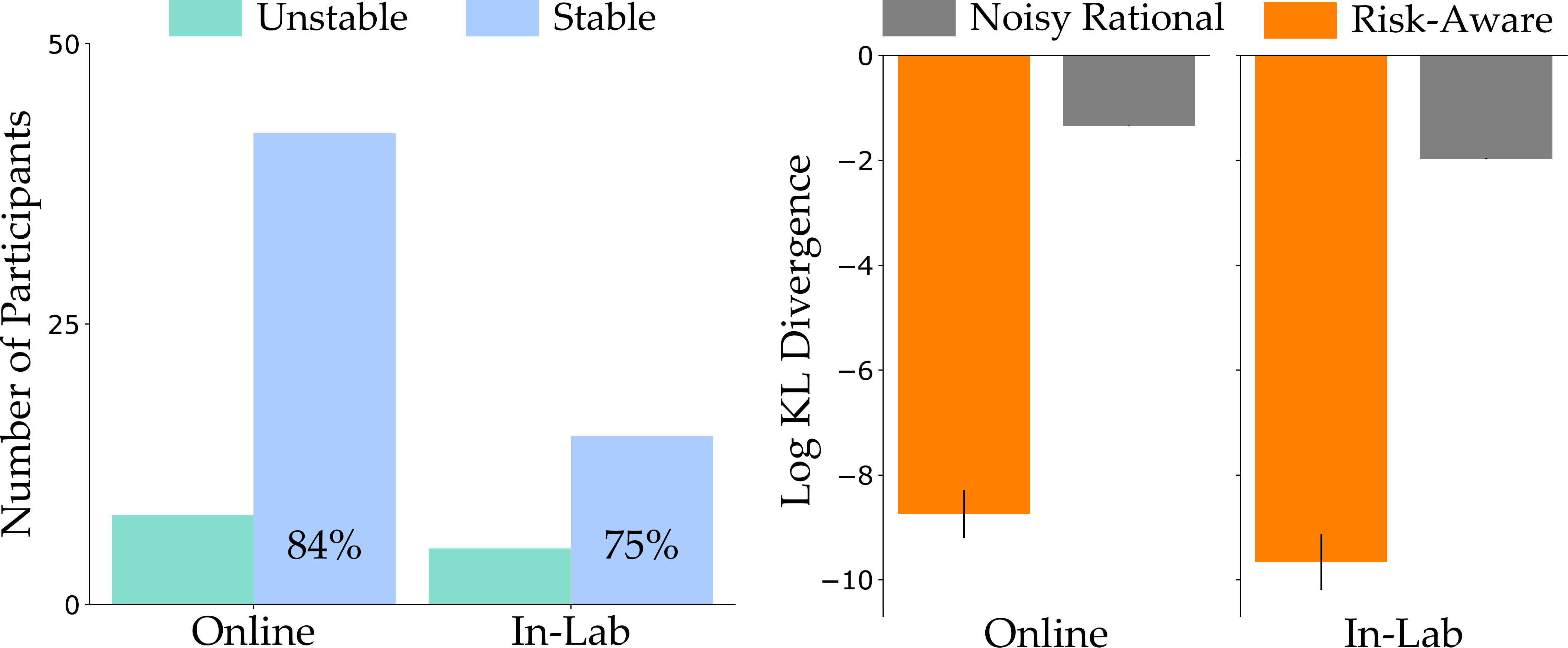}

		\caption{Results from online and in-person user studies during the collaborative cup stacking task. (Left) Although building the unstable tower was \textit{optimal}, more participants selected the stable tower. (Right) Model accuracy, where lower is better. The \textit{Risk-Aware} robot was better able to predict which cups the human would pick up.}
		\label{fig:robot_KL}
	\end{center}
	\vspace{-1.5em}
\end{figure}

\begin{figure*}
    \centering
    \includegraphics[width=1.8\columnwidth]{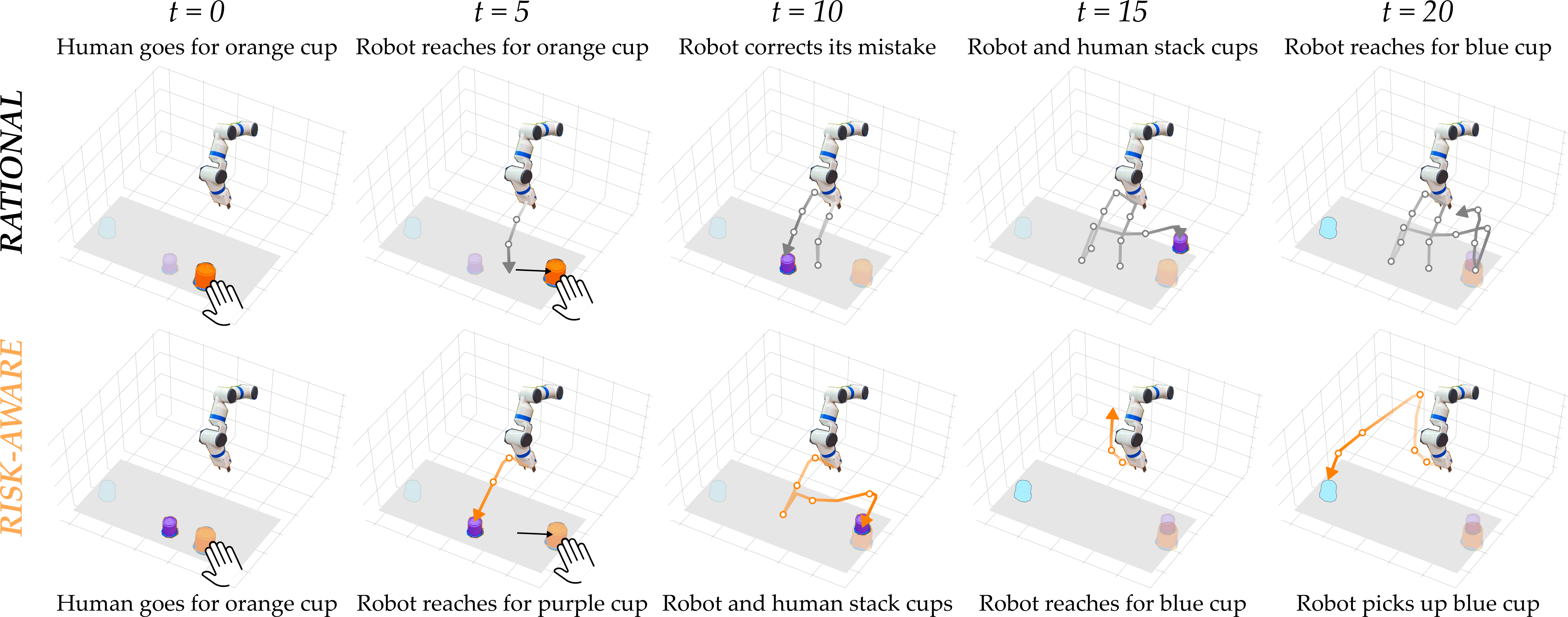}

    \caption{Example robot and human behavior during the collaborative cup stacking user study. At the start of the task, the human reaches for the orange cup (the first step towards a stable tower). When the robot models the human as a \textit{Noisy Rational} partner (top row), it incorrectly anticipates that the human will build the optimal but unstable tower; this leads to interference, replanning, and a delay. The robot leveraging our \textit{Risk-Aware} formalism (bottom row) understands that real decisions are influenced by uncertainty and risk, and correctly predicts that the human wants to build a stable tower. This results in safer and more efficient interaction, leading to faster tower construction.}
    \label{fig:time_series}
\end{figure*}

\prg{Results} Our results from the \textbf{online} user study are summarized in \figref{fig:robot_KL}. During this scenario---where the human is collaborating with the robot---we observed a bias towards risk-averse behavior. Participants overwhelmingly preferred to build the stable tower (and take the guaranteed reward), even though this choice was suboptimal. Only the \textit{Risk-Aware} robot was able to capture and predict this behavior: inspecting the right side of \figref{fig:robot_KL}, we found a statistically significant improvement in model \textit{accuracy} across the board ($t(59) = -21.1$, $p<.001$). Focusing only on the \textbf{online} users, the \textit{log KL divergence} for \textit{Risk-Aware} reached $-8.7 \pm 3.0$, while \textit{Noisy Rational} remained at $-1.3 \pm 0.01$ ($t(29) = -13.1$, $p<.001$). Overall, these results match our findings from the autonomous driving domain, and support hypothesis H2.

\smallskip

\subsection{Planning with Risk-Aware Human Models}

\smallskip

Having established that the \textit{Risk-Aware} robot more accurately models the human's actions, we next explored whether this difference is \textit{meaningful} in practice. We performed an \textbf{in-lab} user study comparing \textit{Noisy Rational} and \textit{Risk-Aware} collaborative robots. We focused on how robots can leverage the \textit{Risk-Aware} human model to improve \textit{safety} and \textit{efficiency} during collaboration.

\smallskip

\noindent\textbf{Participants and Procedure.} Ten members of the Stanford University community ($2$ female, ages $20 - 36$) provided informed consent and participated in this study. Six of these ten had prior experience interacting with the Fetch robot. We used the same experimental setup, rewards, and probabilities described at the beginning of the section. Participants were encouraged to build towers to maximize the total number of points that they earned.

Each participant had ten familiarization trials to practice building towers with the robot. During these trials, users learned about the probabilities of each type of tower collapsing from experience. In half of the familiarization trials, the robot modeled the human with the \textit{Noisy Rational} model, and in the rest the robot used the \textit{Risk-Aware} model; we randomly interspersed trials with each model. After the ten familiarization trials, users built the tower once with \textit{Noisy Rational} and once with \textit{Risk-Aware}: we recorded their choices and the robot's performance during these final trials. The presentation order for these final two trials was counterbalanced.

\smallskip

\noindent\textbf{Dependent Measures.} To test \textit{efficiency}, we measured the time taken to build the tower (Completion Time). We also recorded the Cartesian distance that the robot's end-effector moved during the task (Trajectory Length). Because the robot had to replan longer trajectories when it interfered with the human, Trajectory Length was an indicator of \textit{safety}.

After participants completed the task with each type of robot (\textit{Noisy Rational} and \textit{Risk-Aware}) we administered a $7$-point Likert scale survey. Questions on the survey focused on four scales: how enjoyable the interaction was (Enjoy), how well the robot understood human behavior (Understood), how accurately the robot predicted which cups they would stack (Predict), and how efficient users perceived the robot to be (Efficient). We also asked participants which type of robot they would rather work with (Prefer) and which robot better anticipated their behavior (Accurate).

\smallskip

\noindent\textbf{Hypotheses.}
\begin{quote}
    \textbf{H3.} \textit{Users interacting with the Risk-Aware robot will complete the task more safely and efficiently.}
\end{quote}
\begin{quote}
    \textbf{H4.} \textit{Users will subjectively perceive the Risk-Aware robot as a better partner who accurately predicts their decisions and avoids grabbing their intended cup.}
\end{quote}

\begin{figure}[t]
	\begin{center}
		\includegraphics[scale=0.32]{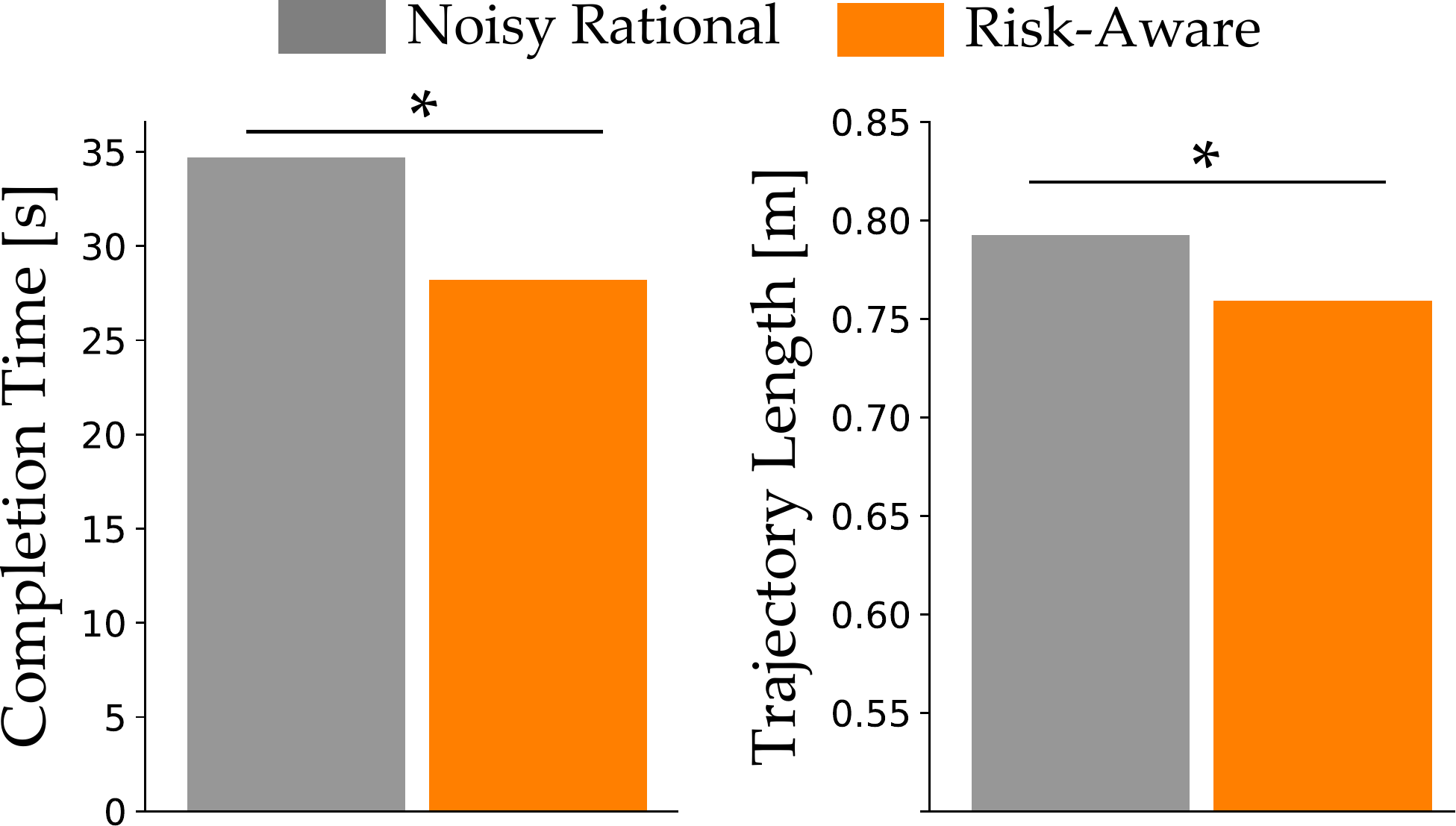}
		\caption{Objective results from our in-lab user study. When participants built the tower with the \textit{Risk-Aware} robot, they completed the task more efficiently (lower Completion Time) and safely (lower Trajectory Length). Asterisks denote significance ($p < .05)$.}
		\label{fig:objective}
	\end{center}
	\vspace{-1em}
\end{figure}

\begin{figure}[t]
	\begin{center}
        \includegraphics[scale=0.45]{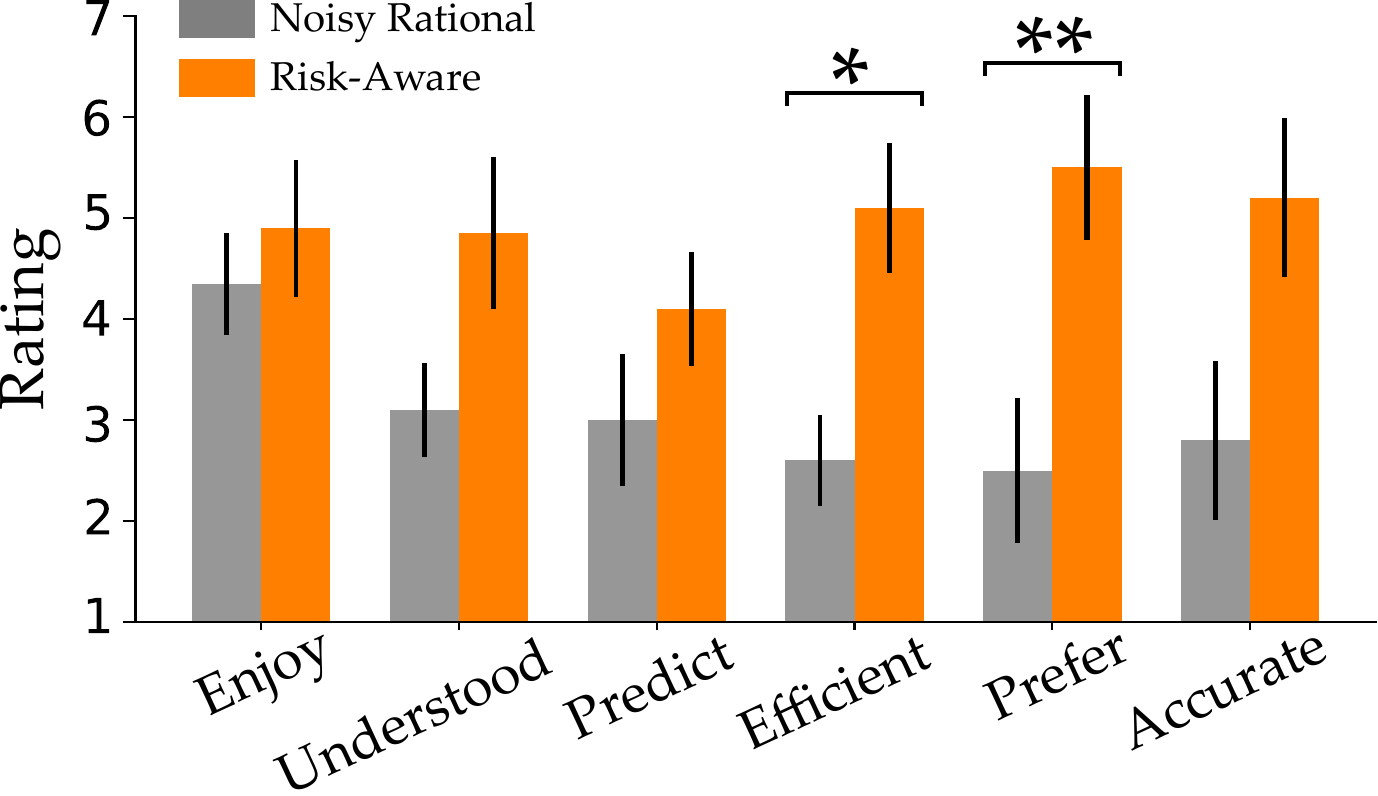}
		\caption{Subjective results from our in-person user study. Higher ratings indicate agreement (i.e., more enjoyable, better understood). Here $**$ denotes $p<.07$, and $*$ denotes $p<.05$. Participants perceived the \textit{Risk-Aware} robot as a more efficient teammate, and marginally preferred collaborating with the \textit{Risk-Aware} robot.}
		\label{fig:subjective}
	\end{center}
	\vspace{-1.744em}
\end{figure}
\noindent\textbf{Results - Objective.} We show example human and robot behavior during the \textbf{in-lab} collaborative cup stacking task in \figref{fig:time_series}. When modeling the human as \textit{Noisy Rational}, the robot initially moved to grab the optimal cup and build the unstable tower. But in $75\%$ of trials participants built the suboptimal but \textit{stable} tower! Hence, the \textit{Noisy Rational} robot often interfered with the human's actions. By contrast, the \textit{Risk-Aware} robot was collaborative: it correctly predicted that the human would choose the stable tower, and reached for the cup that best helped build this tower. This led to improved safety and efficiency during interaction, as shown in \figref{fig:objective}. Users interacting with the risk-aware robot completed the task in less time ($t(9)=2.89$, $p<.05$), and the robot partner also traveled a shorter distance with less human interference ($t(9)=2.24$, $p<.05$). These objective results support hypothesis H3.


\noindent\textbf{Results - Subjective.} We plot the user's responses to our $7$-point surveys in \figref{fig:subjective}. We first confirmed that each of our scales (\textit{Enjoy}, \textit{Understood}, etc.) was consistent, with a Cronbach's alpha $> 0.9$. We found that participants marginally preferred interacting with the \textit{Risk-Aware} robot over the \textit{Noisy Rational} one ($t(9)=2.09$, $p<.07$). Participants also indicated that they felt that they completed the task more efficiently with the \textit{Risk-Aware} robot ($t(9)=3.01$, $p<.05$). The other scales favored \textit{Risk-Aware}, but were not statistically significantly. Within their comments, participants noticed that the \textit{Noisy Rational} robot clashed with their intention: for instance, \textit{``it tried to pick up the cup I wanted to grab''}, and \textit{``the robot picked the same action as me, which increased time''}. Overall, these subjective results partially support hypothesis H4.


\noindent\textbf{Summary.} Viewed together, our \textbf{online} and \textbf{in-lab} user studies \textit{not only} extended our autonomous driving results to a collaborative human-robot domain, but they \textit{also} demonstrated how robots can leverage our formalism to meaningfully adjust their behavior and improve safety and efficiency. Our in-lab user study showed that participants interacting with a \textit{Risk-Aware} robot completed the task faster and with less interference. We are excited that robots can actively use their \textit{Risk-Aware} model to improve collaboration.
\section{Discussion and Conclusion}
Many of today's robots model human partners as Noisy Rational agents. In real-life scenarios, however, humans must make choices subject to uncertainty and risk---and within these realistic settings, humans display a cognitive bias towards \emph{suboptimal} behavior. We adopted Cumulative Prospect Theory from behavioral economics and formalized a human decision-making model so that robots can now anticipate suboptimal human behavior. Across autonomous driving and collaborative cup stacking environments, we found that our formalism better predicted user decisions under uncertainty. We also leveraged this prediction within the robot's planning framework to improve \emph{safety} and \emph{efficiency} during collaboration: our Risk-Aware robot interfered with the participants less and received higher subjective scores than the Noisy Rational baseline. We want to emphasize that this approach is \textit{different from making robots robust} to human mistakes by \textit{always} acting in a risk-averse way. Instead, when humans prefer to take safer but suboptimal actions, robots leveraging our formalism \textit{understand} these conservative humans and increase overall team performance.


\noindent \textbf{Limitations and Future Work.} A strength and limitation of our approach is that the Risk-Aware model introduces additional parameters to the state-of-the-art Noisy Rational human model. With these additional parameters, robots are able to predict and plan around suboptimal human behavior; but if not enough data is available when the robot learns its human model, the robot could overfit. We point out that for all of the user studies we presented, the robots learned Noisy Rational and Risk-Aware models from the \emph{same amount} of user data. 

When learning and leveraging these models, the robot must also have access to real-world information. Specifically, the robot must know the rewards and probabilities associated with the human's decision. We believe that robots can often obtain this information from experience: for example, in our collaborative cup stacking task, the robot can determine the likelihood of the unstable tower falling based on previous trials. Future work must consider situations where this information is not readily available, so that the robot can identify collaborative actions that are \emph{robust} to errors or uncertainty in the human model.

Finally, we only tested the Risk-Aware model in bandit settings where the horizon is $1$. Ideally, we would want our robots to be able to model humans over longer horizons. We attempt to address part of this limitation by conducting a series of experiments in a grid world setting with a longer horizon. We found that a Risk-Aware robot can more accurately model a sequence of human actions as compared to the Noisy Rational robot. Experiment details and results are further explained in the Appendix. 

Collaborative robots need algorithms that can predict and plan around human actions in real world scenarios. We proposed an extension of Noisy Rational human models that also accounts for suboptimal decisions influenced by risk and uncertainty. While user studies across autonomous driving and collaborate cup stacking suggest that this formalism improves model accuracy and interaction safety, it is only one step towards seamless collaboration.
\begin{acks}
Toyota Research Institute ("TRI")  provided funds to assist the authors with their research but this article solely reflects the opinions and conclusions of its authors and not TRI or any other Toyota entity.
\end{acks}

\bibliographystyle{ACM-Reference-Format}
\bibliography{sample-sigconf.bib}



\appendix
\section*{Appendix}
To investigate how well \textit{Risk-Aware} and \textit{Noisy Rational} model humans in more complex POMDP settings, we designed two different maze games. Each game consists of two $17$-by-$15$ grids and these two grids have the exact same structure of walls, which are visible to the player. In each grid, there is one \emph{start} and two \emph{goal} squares. Players start from the same square, and reach either of the goals. Each square in the grids has an associated reward, which the player can also observe. The partial observability comes from the rule that the player does not exactly know which grid she is actually playing at. While she is in the first grid with $95\%$ probability, there is a $5\%$ chance that she might be playing in the second grid. We visualize the grids for both games in \figref{fig:compact_mazes}, and also attach the full mazes in the supplementary material. We restricted the number of moves in each game such that the player has to go to the goals with the minimum possible number of moves. Finally, we enforced a time limit of $2$ minutes per game.

\begin{figure}[h]
	\includegraphics[width=\columnwidth]{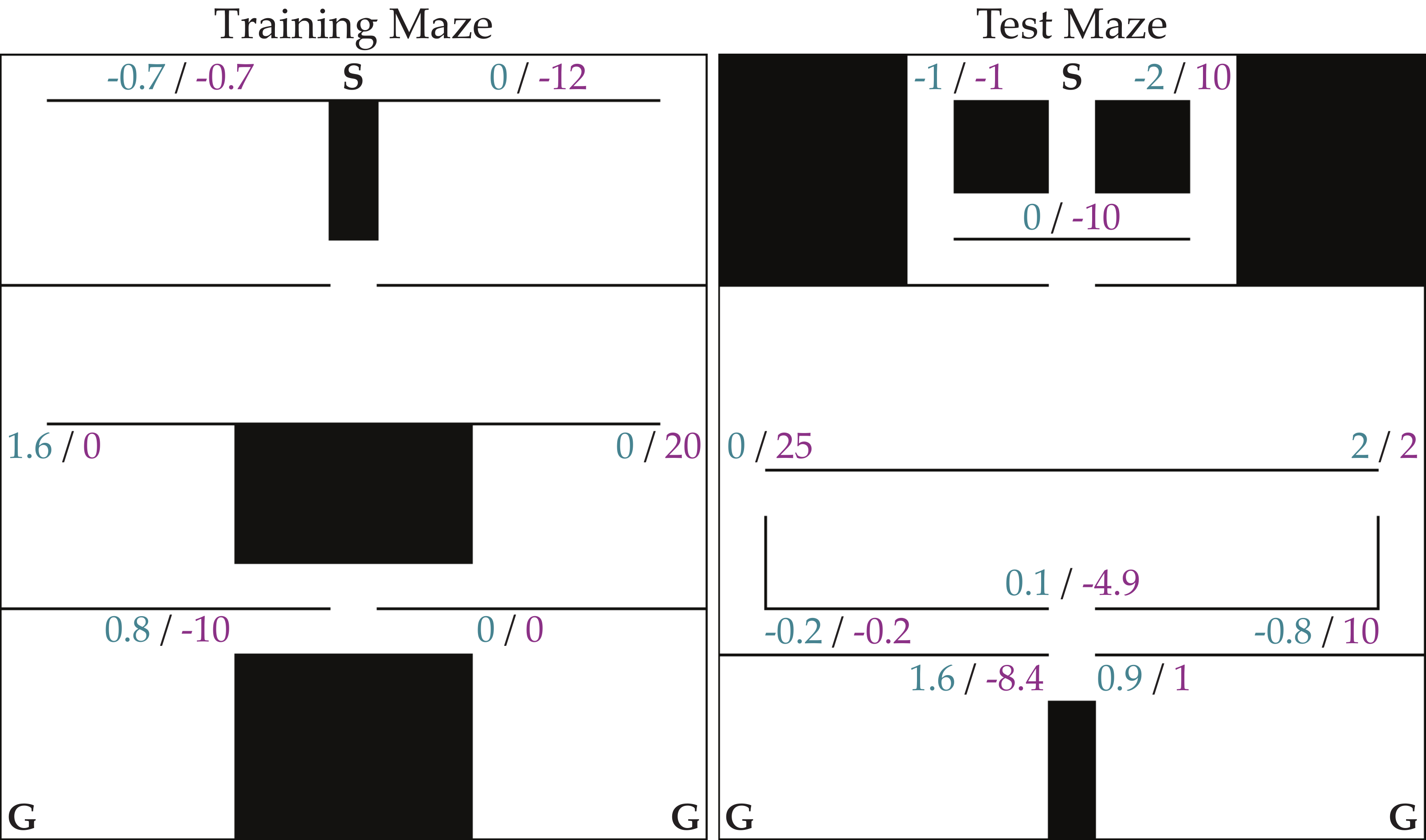}
	\caption{Summaries of two games. For each game, we have a maze. The values written on the mazes represent how much reward players can collect by entering those roads. The first numbers in each pair correspond to the $95\%$ grid, and the second one to the $5\%$ grid.}
	\label{fig:compact_mazes}
\end{figure}

We investigate the effect of both risk and time constraints via this experiment. While it is technically possible for the players to compute the optimal trajectory that leads to the highest expected reward, time limitation makes it very challenging, and humans resort to rough calculations and heuristics. Moreover, we designed the mazes such that humans can get high rewards or penalties if they are in the low-probability ($5\%$) grid. This helps us investigate when humans become risk-seeking or risk-averse.

We recruited 17 users (4 female, 13 male, median age 23), who played both games. We used one game (two grids) to fit the model parameters independently for each user, and the other game (other two grids) to evaluate how well the models can explain the human behavior. As the human actions depend not only the immediate rewards, but also the future rewards, we ran value iteration over the grids and used the values to fit the models as we described in Sec.~3. We again employed Metropolis-Hastings to sample model parameters, and recorded the mean of the samples.

\begin{figure}[h]
	\includegraphics[width=\columnwidth]{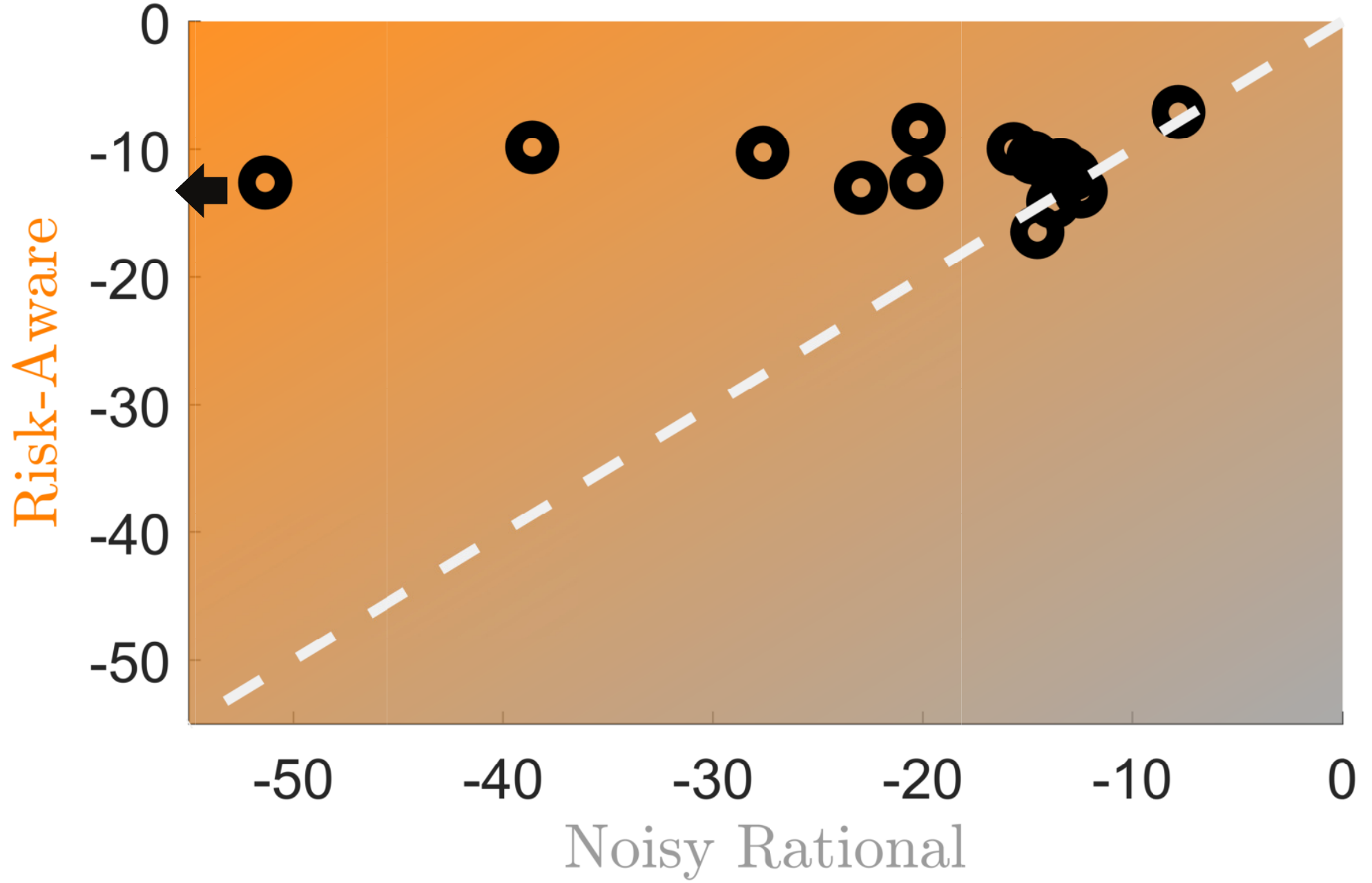}
	\caption{Log-Likelihood values by Risk-Aware and Noisy Rational models. One outlier point is excluded from the plot and is shown with an arrow.}
	\label{fig:maze_results}
\end{figure}

Figure~\ref{fig:maze_results} shows the log-likelihoods for each individual user for Risk-Aware and Noisy Rational models. Overall, Risk-Aware explains the test trajectories better. The difference is statistically significant (paired $t$-test, $p<0.05$). In many cases, we have seen risk-averse and risk-seeking behavior from people. For example, $12$ out of of $17$ users chose the risk-seeking action in the test maze by trying to get $25$ reward with probability $5\%$ instead of getting $2$ with $100\%$ probability. Similarly, $15$ out of $17$ users choose to guarantee $0.9$ reward and gain $0.1$ more with $5\%$ probability instead of guaranteeing $1.6$ reward and losing $10$ with $5\%$ probability. This is an example of suboptimal risk-averse action.


\end{document}